\documentclass[11pt]{article}
\usepackage[margin=1in]{geometry}
\usepackage[T1]{fontenc}
\usepackage[utf8]{inputenc}
\usepackage{newtxtext}
\usepackage{newtxmath}
\usepackage{bm}
\usepackage{amsmath}
\usepackage{setspace}
\newcommand{\SingleSpacedXII}{\singlespacing}
\usepackage{algorithm}
\usepackage{algpseudocode}
\usepackage{tikz}
\usepackage{url}
\usepackage{natbib}
\bibpunct[, ]{(}{)}{,}{a}{}{,}%
\def\bibsep{\smallskipamount}%
\usepackage{graphicx}
\usepackage{booktabs}
\usepackage{pdfpages}
\usepackage{etoolbox}
\apptocmd{\normalsize}{%
	\setlength{\abovedisplayskip}{3pt}%
	\setlength{\belowdisplayskip}{3pt}%
	\setlength{\abovedisplayshortskip}{0pt}%
	\setlength{\belowdisplayshortskip}{2pt}%
}{}{}
\usepackage{xcolor}
\definecolor{linkblue}{HTML}{1b4081}
\usepackage{caption}
\DeclareCaptionLabelFormat{boldblue}{\textcolor{linkblue}{\textbf{#1 #2.}}}
\usepackage{hyperref}
\hypersetup{
	colorlinks=true,
	linkcolor=linkblue,
	citecolor=linkblue,
	urlcolor=cyan,
	pdftitle={Deconstructing Intraocular Pressure: A Non-Invasive Multi-Stage Probabilistic Inverse Framework},
	pdfauthor={Md Rezwan Jaher, Abul Mukid Mohammad Mukaddes, A. B. M. Abdul Malek}
}

\title{Deconstructing Intraocular Pressure: A Non-Invasive Multi-Stage Probabilistic Inverse Framework}
\author{Md Rezwan Jaher\thanks{Department of Mathematics and Physics, North South University, Dhaka 1229, Bangladesh. Email: \href{mailto:rezwan.plabon@gmail.com}{rezwan.plabon@gmail.com}. ORCID: \href{https://orcid.org/0009-0002-1692-5967}{0009-0002-1692-5967}. Corresponding author. Part of this work was done while the author was an independent researcher.}
\and Abul Mukid Mohammad Mukaddes\thanks{Department of Industrial \& Production Engineering, Shahjalal University of Science and Technology, Sylhet, Bangladesh. Email: \href{mailto:mukaddes-ipe@sust.edu}{mukaddes-ipe@sust.edu}.}
\and A. B. M. Abdul Malek\thanks{Department of Industrial \& Production Engineering, Shahjalal University of Science and Technology, Sylhet, Bangladesh. Email: \href{mailto:bashar@sust.edu}{bashar@sust.edu}.}}
\date{}

\begin{document}
\maketitle

\begin{abstract}
Here, we propose the \textit{mechanistically hierarchical} inverse problem (MHIP) paradigm to address a distinct class of challenges in biological transport systems, where a macroscopic observation emerges from the multiplicative entanglement of intrinsic micro-scale physics and latent macro-scale geometry. Due to a fundamental lack of \textit{in vivo} ground truth, solving MHIPs from clinical data is severely underdetermined. Consequently, standard statistical sampling methods suffer from pathological degeneracy along non-identifiable manifolds, while monolithic machine learning baselines collapse to an uninterpretable conditional mean.

To overcome this extreme ill-posedness, we propose a two-stage physics-guided architectural decomposition that transforms a globally non-identifiable inverse mapping into conditionally identifiable sub-problems. The first stage acts as a constrained inverse surrogate, isolating the intrinsic biophysical parameter under a fixed reference manifold. The second stage performs empirical structural calibration, bridging the idealized physics with macroscopic reality. We also introduce Physics-Calibrated Data Scaling (PCDS), a tailored multi-fidelity modeling strategy that operates under extreme data scarcity to render MHIPs tractable without requiring patient-specific 3D simulations.

We demonstrate the utility of this framework through a critical clinical case study in glaucoma, computationally deconstructing a single intraocular pressure (IOP) measurement to estimate both its unmeasurable biophysical root cause (tissue permeability, $\hat{K}_{\text{TM}}$) and the clinical gold standard for outflow function ($\hat{C}_{\text{trab}}$). In retrospective validation against published \textit{in vivo} data, the non-invasive $\hat{C}_{\text{trab}}$ estimates showed low bias against measured outflow facility. Ablation studies support the diagnostic value of our $\hat{K}_{\text{TM}}$ estimate and the benefit of architectural decoupling. While naive monolithic baselines achieve only 37.0\% diagnostic accuracy, incorporating the isolated $\hat{K}_{\text{TM}}$ biomarker boosts performance to 85.2\%. Ultimately, the full two-stage architecture reaches 88.9\% accuracy while generating a probabilistic profile that maps the complex interplay of biophysical root causes, offering a generalizable computational blueprint for solving MHIPs.
\end{abstract}


\section{Introduction}
\label{sec:Intro}
In computational science, inverse problems are fundamental for inferring latent parameters from indirect observations \citep{Tarantola2005, Stuart2010, Nelsen2025}. While ``hierarchical inverse problems'' typically denote \textit{statistical} hierarchies (e.g., nested probability distributions) \citep{Agapiou2014, Latz2019} or \textit{computational} nesting (e.g., algorithmic sub-procedures) \citep{Oughton2016, GramHansen2019}, we introduce a fundamentally different form: \textit{mechanistically} hierarchical inverse problems (MHIPs). In these systems, an observable macroscopic quantity emerges from the multiplicative entanglement of intrinsic micro-scale physics and latent macro-scale geometry. For example, under Darcy's law for porous media, an observed pressure gradient is the entangled product of intrinsic material permeability and macroscopic spatial dimensions. In biological transport systems, MHIPs are severely underdetermined due to two limitations. First, direct \textit{in vivo} measurement of micro-scale properties is clinically infeasible, precluding ground-truth data. Second, the multiplicative entanglement of governing physics creates severe structural non-identifiability. Because diverse physiological parameter combinations yield the exact same macroscopic observation, the inverse mapping violates Hadamard's requirement for solution uniqueness \citep{Hadamard1923, Tarantola2005}. Nevertheless, solving these systems is of utmost clinical importance to identify the hidden biophysical root causes of observable symptoms. 

To overcome these challenges, we present a two-stage framework integrating a physics-guided architectural decomposition with a calibrated synthetic data generation strategy. We explore its formulation and utility through a clinical case study of ocular hydrodynamics and glaucoma.

\subsection{Motivation and Background: The Glaucoma Case Study}

The biological transport of aqueous humor exemplifies a mechanistically hierarchical system. Glaucoma, a leading cause of blindness \citep{RN1,RN2}, is driven by elevated intraocular pressure (IOP) \citep{RN3}. Under Darcy's Law, IOP emerges from the multiplicative entanglement of intrinsic trabecular meshwork (TM) permeability ($K_{\text{TM}}$) \citep{RN6,RN7,RN8} and macroscopic geometric efficiency ($A/L$). Clinical diagnosis requires solving the \textit{inverse} problem: deducing the unmeasurable micro-scale root cause ($K_{\text{TM}}$) from the macroscopic consequence (IOP). However, historical computational models remain deterministic and forward-focused \citep{RN38, RN45, Sigal2005, RN42}. By requiring $K_{\text{TM}}$ as a fixed input to predict IOP, they create a circular dependency ignoring patient-specific variability in latent boundary hydrodynamics like uveoscleral outflow ($F_u$) and episcleral venous pressure (EVP). Consequently, mapping a sparse clinical observation (IOP) and demographic metadata back to a functional metric like conventional outflow facility (OF), $C_{\text{trab}}$ \citep{RN10} constitutes a severely underdetermined MHIP. Expanded details on the ocular anatomy, current clinical measurement gaps, and the complete methodological pipeline are provided in \textbf{Online Supplement Note 1 (Figure S1 and S2)}.

\subsection{Related Work and Computational Paradigms for Inverse Problems}
While successful in their intended domains, existing computational strategies face fundamental limitations under the extreme data scarcity and structural non-identifiability characteristic of biological transport MHIPs.
\subsubsection{Prior Inverse Approaches in Ocular Biomechanics and Hydrodynamics}
Recent inverse modeling in ophthalmology remains restricted by stringent data requirements and an inability to directly resolve hydrodynamics. For instance, inverse FE methods coupled with OCT estimate latent biomechanical properties \citep{Girard2013, Pant2017}, but require rich spatial displacement fields mapped during complex \textit{in vivo} loading (e.g., active indentation). This renders them clinically impractical for routine screening. Furthermore, these elastography-based solvers map only solid mechanics, leaving critical fluid resistance determinants, such as hydraulic permeability ($K_{\text{TM}}$) and OF ($C_{\text{trab}}$), unresolved. Similarly, formal Bayesian inference applied to ocular transport \citep{Crawshaw2025} relies on dense, longitudinal time-series data for parameter identifiability, rendering monolithic statistical calibrations unusable for single-snapshot clinical assessments.
\subsubsection{Statistical and Algebraic Approaches to Hierarchical Inverse Problems}
\label{sec:statistical&algebraic}
The foundational Bayesian framework \citep{Stuart2010, SanzAlonso2023}, alongside statistical calibration, metamodeling, and hierarchical emulation \citep{Kennedy2001, Barton2006, Oughton2016}, excels in computationally nested systems but encounters severe identifiability challenges in MHIPs. Constructing a joint posterior over multiplicatively entangled parameters from sparse scalar observations is structurally confounded. The likelihood forms a continuous, non-identifiable manifold. This manifold is a hyperbolic surface where the parameter product remains constant for any fixed observation \citep{Tarantola2005}. Consequently, MCMC algorithms suffer from pathological mixing \citep{Agapiou2014}, and modern amortized posteriors \citep{Arridge2019} merely mirror this uninformative ridge geometry rather than resolving the non-identifiability. Thus, beyond the steep computational cost of sequential evaluations \citep{Nelsen2025}, monolithic Bayesian approaches require explicit architectural decoupling to isolate micro-scale physics from macro-scale variance. Similarly, algebraic OR hierarchies \citep{Gribanova2022} efficiently manage closed-form networks but treat hierarchical factors as abstract nodes. They cannot natively embed coupled partial differential equations (PDEs), facing fundamental representational challenges when macroscopic outputs emerge from continuous micro-scale vector fields rather than simple algebraic arithmetic.
\subsubsection{Physics-Informed Neural Networks (PINNs)}
To bypass MCMC bottlenecks, deep neural networks \citep{Arridge2019}, specifically PINNs \citep{Raissi2019}, embed PDE residuals into the loss function. However, standard PINNs strictly require a defined spatial domain for collocation point sampling. In biological MHIPs, physical geometry ($\gamma$) is typically a latent unknown (e.g., high-fidelity \textit{in vivo} imaging of trabecular micro-architecture remains clinically infeasible). Without spatial ground truth, the computational domain is undefined and the physics loss cannot be evaluated. While geometry-agnostic variants like physics-informed operator networks \citep{Wang2021} address this rigidity, they still require dense functional measurements for identifiability \citep{Molinaro2023}. Given only a single scalar observation, even an approximately parameterized latent geometry cannot resolve the multiplicative entanglement with the intrinsic parameter ($u$). The physics-constrained loss remains non-identifiable along the exact same hyperbolic ridge (Section~\ref{sec:statistical&algebraic}), dictating that the optimization landscape inevitably collapses to an uninterpretable mean without explicit architectural decoupling.

\subsubsection{Monolithic Operator Learning and Simulation-Guided Supervised Learning}
Deep neural operators and simulation-guided models successfully learn forward and inverse mappings from synthetic data \citep{Kovachki2023, Lu2021}, extending even to compositionally chained multiphysics operators \citep{Yin2023} and end-to-end neural inverse operators \citep{Nelsen2025}. However, while effective for well-posed tasks, these supervised approaches require paired input-output data. Because intrinsic constitutive properties ($K_{\text{TM}}$) are unmeasurable \textit{in situ}, compiling real-world \textit{in vivo} training datasets is infeasible. Relying instead on simulation-guided synthetic data forces the architecture to map a sparse scalar observation back to a multi-scale physical state. Since varying combinations of intrinsic permeability and macroscopic geometry yield identical macroscopic outputs, this inverse mapping remains fundamentally non-injective. 

Faced with this severe structural non-identifiability, monolithic architectures \citep{Molinaro2023, Yin2023} treat the underlying physical hierarchy as a black box. Consequently, deterministic operator networks and simulation-guided regressors inherently conflate structural variance and intrinsic physics, mathematically collapsing multi-modal conditional distributions into a single, blurry conditional mean \citep{Nelsen2025}. Theoretically, this forces predictions to exhibit severely suppressed variance, clustering tightly around uninformative population averages regardless of instance-specific inputs. As empirically demonstrated in our ablation study (see \textbf{Online Supplement Note 6} for full details), this inability to isolate specific micro-scale mechanisms renders monolithic models structurally unsuited for instance-specific mechanistic risk stratification.
\subsection{The Gap Addressed and Summary of Contributions}
\label{sec:gap_and_contributions}
The limitations of the aforementioned paradigms highlight a critical gap: traditional Bayesian calibration is prone to degeneracy under multiplicative entanglement; algebraic OR hierarchies lack the capacity to embed coupled continuum physics; PINNs depend on fully observed spatial domains; and monolithic supervised ML collapses to an uninterpretable conditional mean. To address this, we formalize the MHIP as a distinct problem class. Distinguished from prior computationally \citep{Oughton2016, GramHansen2019} or compositionally \citep{Yin2023} nested models, we exploit the physical origin of the hierarchy rather than treating it as a black box. Instantiating this paradigm through a clinical case study of glaucoma, we tackle a complex biological system severely confounded by a fundamental lack of \textit{in vivo} ground truth. We computationally deconstruct a single routine intraocular pressure (IOP) observation to estimate its unmeasurable biophysical root cause (trabecular permeability, $K_{\text{TM}}$) and the clinical gold standard for outflow function ($C_{\text{trab}}$). Our key methodological and applied contributions are as follows:

\begin{itemize}
	
	\item \textbf{Architectural Paradigm for MHIPs:} To our knowledge, we present the first formalization of multiplicative physical entanglement as a \textit{mechanistic hierarchy}. Rather than treating this entanglement as an obstacle, we exploit it as structural information. We propose a physics-guided two-stage decomposition that isolates the intrinsic parameter under a fixed reference manifold before applying empirical structural calibration. This transforms a globally non-identifiable inverse problem into conditionally identifiable mappings, structurally mitigating the predictive collapse inherent to monolithic architectures.
	
	\item \textbf{Scalable Data Generation via PCDS:} We introduce PCDS, a domain-adapted extension of multi-fidelity delta learning \citep{Peherstorfer2018}. By combining clinical anchoring, noise-injection regularization, and structural error-compensation, PCDS generates 120,000 physically consistent training samples from only 500 targeted high-fidelity simulations. The full dataset was generated in approximately eight hours.
	
	\item \textbf{Novel Functional Biomarker and Clinical Validation:} To our knowledge, our framework provides the first non-invasive computational estimates of both a functional trabecular permeability biomarker ($\hat{K}_{\text{TM}}$) and the current clinical gold standard for outflow function ($\hat{C}_{\text{trab}}$) from sparse routine inputs. In retrospective validation, $\hat{K}_{\text{TM}}$ tracks measured outflow facility by rank. $\hat{C}_{\text{trab}}$ agrees with measured outflow facility with low bias.
	
	\item \textbf{Probabilistic Profiling and Risk Stratification:} Constrained Monte Carlo sampling generates empirical push-forward distributions over all latent hydrodynamic states. An ablation across 27 cohorts supports the architecture. Naive monolithic baselines achieve only 37.0\% diagnostic accuracy. Providing our isolated $\hat{K}_{\text{TM}}$ biomarker raises monolithic accuracy to 85.2\%. The full two-stage framework reaches 88.9\% when cohorts are classified by $\hat{C}_{\text{trab}}$. Classifying directly by $\hat{K}_{\text{TM}}$ reaches 92.6\% (25 of 27 cohorts). These results suggest diagnostic value in separating intrinsic physics from geometric variation.
\end{itemize}

\section{Methods}
\label{sec:methods}
Here, we provide the detailed methodology for the development and validation of our physics-guided, multi-stage probabilistic framework. We first establish the rigorous mathematical formulation of the inverse problem and its associated error taxonomy. Next, we present the physical rationale and architecture of our multi-scale framework, translating these abstract concepts into the biophysics of ocular hydrodynamics and detailing our Physics-Calibrated Data Scaling (PCDS) strategy. We then describe the computational model development used to generate the foundational training data, including the FE geometry, governing equations, and numerical implementation. This is followed by a description of the supervised learning tasks, the clinical and \textit{in silico} cohorts curated for validation, and the statistical methods employed for all analyses.

\subsection{Mathematical Formulation and Error Taxonomy}
\label{sec:math_formulation}

We define a \textit{MHIP} as one where the forward mapping is an explicit composition of physical laws operating across multiple scales. Rather than a monolithic black-box mapping $y = \mathcal{F}(u) + \eta$, the observable Quantity of Interest (QoI, $y$) is the output of nested physical operators:
\begin{equation}
	y = \mathcal{F}_{\text{macro}} \big( \mathcal{F}_{\text{micro}}(u, \gamma), \mathbf{x} \big) + \eta
	\label{eq:structural_hierarchy}
\end{equation}
where $\eta$ denotes observation noise, written $\epsilon_{\text{noise}}$ below. We assume finite-dimensional parameter spaces $\mathcal{U} \subset \mathbb{R}^k$, $\Gamma \subset \mathbb{R}^l$ and an observation space $\mathcal{Y} \subset \mathbb{R}^m$. To ensure conditional identifiability when isolating the intrinsic parameter, we require $k \le m$, and assume the forward micro-scale operator $\mathcal{F}_{\text{micro}}$ is strictly monotonic with respect to $u$ under fixed boundaries (a standard property of mass and momentum conservation laws). In our motivating clinical application, the observation is a single scalar measurement (IOP) and the intrinsic parameter is a scalar (permeability), making $k=m=1$. This framework is suited to systems where:
\begin{enumerate}
	\item The forward map decomposes compositionally according to known physical laws.
	\item One component ($u \in \mathcal{U}$) governs intrinsic, constitutive micro-scale physics.
	\item Another component ($\gamma \in \Gamma$) governs macro-scale structure or geometry.
	\item The observation dimension is severely lower than the latent joint parameter dimension ($m \ll k+l$), but sufficient for the isolated intrinsic parameter ($k \le m$).
	\item Direct \textit{in vivo} or \textit{in situ} measurement of the structural component $\gamma$ is infeasible.
\end{enumerate}

For many transport systems, a reduced representation takes the form $y \approx \phi(u)\psi(\gamma)$, meaning the observable QoI depends on the multiplicative entanglement of intrinsic root causes and structural factors. Consequently, this entanglement renders the inverse problem \textit{structurally non-identifiable} when relying on a single scalar observation. To address this extreme ill-posedness without requiring intractable high-fidelity structural simulation or imaging, we propose an architectural decomposition aligned with the mechanistic hierarchy.

\vspace{0.5em}
\noindent \textbf{Stage 1: Constrained Inverse Surrogate under a Fixed Structural Manifold.} \\
In the first stage, we isolate the intrinsic constitutive parameter ($u$) by fixing the macro-scale structure to an idealized reference manifold, $\gamma^{\text{ideal}}$, utilizing high-fidelity numerical simulations. The restricted forward model becomes $y = \mathcal{F}_{\gamma^{\text{ideal}}}(u, \mathbf{x})$. We train a surrogate, $\mathcal{S}_1$, to learn an approximate left-inverse parameterized by this reference manifold:
\begin{equation}
	\hat{u}_{\text{eff}} = \mathcal{S}_{1, \gamma^{\text{ideal}}}(y, \mathbf{x})
\end{equation}
Critically, $\mathcal{F}_{\text{micro}}(u, \gamma)$ is physically independent of the physiological boundary conditions $\mathbf{x}$, as the intrinsic constitutive physics are determined solely by tissue properties and geometry. The boundary conditions $\mathbf{x}$ enter exclusively at the macroscopic level. To learn an approximate left-inverse of this composition, the surrogate $\mathcal{S}_1$ must therefore ingest both $y$ and $\mathbf{x}$ to disentangle the macroscopic confounding, not because $\mathbf{x}$ governs the micro-scale physics, but because traversing the compositional hierarchy in reverse requires knowledge of all variables that participated in the forward direction.

Fixing $\gamma = \gamma^{\text{ideal}}$ converts the joint inversion into a constrained inversion under \textit{model misspecification}. Because the true structure $\gamma^{\text{true}}$ differs from $\gamma^{\text{ideal}}$, the surrogate must satisfy the governing physical laws using the constrained reference. Consequently, the intrinsic parameter estimate becomes an \textit{effective parameter} ($\hat{u}_{\text{eff}}$) containing a systematic, learnable distortion.

\vspace{0.5em}
\noindent \textbf{Stage 2: Empirical Structural Calibration via Population Anchoring.} \\
Stage 2 acts as a structural completion of latent geometric degrees of freedom through population anchoring. To recover the ultimate macroscopic QoI, we introduce a learned operator $\mathcal{C}$:
\begin{equation}
	\hat{\gamma}_{\text{eff}} = \mathcal{C}(\hat{u}_{\text{eff}}, y, \mathbf{x})
\end{equation}
Because the mapping in Stage 1 is a many-to-one surrogate projection, the geometric bias cannot be resolved through simple algebraic inversion without the contextualizing observation $y$. Thus, $\mathcal{C}$ is supervised against structural targets derived from population-anchored distributions of the macroscopic QoI. Importantly, $\hat{\gamma}_{\text{eff}}$ should be interpreted as an effective geometric calibration factor required to reconcile the biased micro-scale estimate with population-level macroscopic targets, rather than a direct reconstruction of patient-specific 3D geometry.

\vspace{0.5em}
\noindent \textbf{Taxonomy of Uncertainties and Error Propagation.} \\
To rigorously formulate our solution, we isolate the distinct sources of uncertainty. Let the true, unobservable state of the physical system be governed by:
\begin{equation}
	y^{\text{true}} = \mathcal{F}_{\text{macro}} \big( \mathcal{F}_{\text{micro}}(u^{\text{true}}, \gamma^{\text{true}}), \mathbf{x}^{\text{true}} \big) 
\end{equation}
When estimating $u$ and $\gamma$ from a clinical observation $y^{\text{obs}} = y^{\text{true}} + \epsilon_{\text{noise}}$, the inference is confounded by three fundamental categories of error. First, the epistemic latency of the true boundary conditions ($\mathbf{x}^{\text{true}}$) is addressed during deployment via \textit{Monte Carlo Uncertainty Propagation}. By drawing $\mathbf{x}^{\text{sampled}} \sim \mathbb{P}(\mathbf{x})$ subject to physiological constraints, the resulting empirical push-forward distribution explicitly maps the parameter variance stemming from unknown boundary conditions.

Second, \textit{Structural Misspecification} arises from the unmeasured geometry $\gamma^{\text{true}}$. The theoretically required intrinsic value is a biased \textit{effective parameter} ($u_{\text{eff}}$), implicitly defining a non-linear distortion mapping $\mathcal{D}_{\text{mismatch}}$:
\begin{equation}
	u_{\text{eff}} = \mathcal{D}_{\text{mismatch}}(u^{\text{true}} ; \gamma^{\text{ideal}}, \gamma^{\text{true}}, \mathbf{x}^{\text{true}})
\end{equation}
Crucially, this structural distortion ($\mathcal{D}_{\text{mismatch}}$) is not arbitrary algorithmic noise; it is a highly structured, deterministic bias. We assume $\mathcal{D}_{\text{mismatch}}$ is a smooth, approximately monotonic function of $u^{\text{true}}$. This assumption is motivated by the mass-conservation and linear momentum structure of the micro-scale law (e.g., Darcy's Law) under fixed boundary conditions. We expect it to hold approximately for the weakly coupled thermal and fluid system solved by our FE model. We have not proven it for that full model. If the bias is structured in this way, the mapping from the effective parameter back to the population-anchored macroscopic target is learnable by the Stage 2 structural calibration operator ($\mathcal{C}$).

Finally, the ML surrogate ($\mathcal{S}_1$) introduces an algorithmic residual ($\epsilon_{\text{ML}}$). Let $u_{\text{eff}}^{\text{obs}}$ denote the theoretical effective parameter that a perfect surrogate would recover to exactly satisfy the mismatched manifold given the noisy inputs ($y^{\text{obs}}, \mathbf{x}^{\text{sampled}}$). Note that $u_{\text{eff}}^{\text{obs}}$ inherently differs from the theoretical noise-free parameter ($u_{\text{eff}}^{\text{true}}$) due to the propagation of measurement noise ($\epsilon_{\text{noise}}$). The surrogate's actual prediction ($\hat{u}_{\text{eff}}$) deviates from this noisy target by a pure algorithmic residual:
\begin{equation}
	\hat{u}_{\text{eff}} = \mathcal{S}_{1,\,\gamma^{\text{ideal}}}(y^{\text{obs}},\, \mathbf{x}^{\text{sampled}}) 
	= u_{\text{eff}}^{\text{obs}} + \epsilon_{\text{ML}}
\end{equation}
Consequently, the total deviation of the Stage 1 prediction from the true intrinsic parameter explicitly decomposes into three distinct mathematical terms:
\begin{equation}
	 \label{eq:error_decomp}
	\hat{u}_{\text{eff}} - u^{\text{true}} = \underbrace{(u_{\text{eff}}^{\text{true}} - u^{\text{true}})}_{\text{structural bias } (\mathcal{D}_{\text{mismatch}})} + \underbrace{(u_{\text{eff}}^{\text{obs}} - u_{\text{eff}}^{\text{true}})}_{\text{propagated input uncertainty}} + \underbrace{\epsilon_{\text{ML}}}_{\text{surrogate error}}
\end{equation}
This explicit decomposition mathematically maps our entire solution architecture. The second term lumps the effects of aleatoric measurement noise (propagated from $y^{\text{obs}}$) and epistemic uncertainty in the latent system inputs (propagated from $\mathbf{x}^{\text{sampled}}$). Both of these propagated uncertainties are explicitly addressed by the architecture: the epistemic uncertainty is marginalized by the Monte Carlo sampling procedure during clinical deployment, while the aleatoric measurement noise is mitigated by the noise-injection regularization during the Stage 2 PCDS training phase (Section~\ref{sec:pcds}). The $\mathcal{S}_1 \to \mathcal{C}$ decomposition is designed to separate the structural bias from these propagated uncertainties. We do not claim that it controls each error term. The resulting distributions are conditional on the assumed input distributions $\mathbb{P}(\mathbf{x})$ and on the trained surrogates.

\subsection{Clinical Instantiation: Ocular Hydrodynamics as an MHIP}
\label{sec:clinical_instantiation}
The abstract MHIP framework formalized in Section~\ref{sec:math_formulation} and Algorithm~\ref{alg:generalized_framework} instantiates directly onto the ocular hydrodynamic system (the complete mathematical-to-clinical variable mapping is provided in \textbf{Online Supplement Table S2}). The observable macroscopic quantity ($y^{\text{obs}}$) is IOP, and the multiplicative entanglement arises from Darcy's Law governing aqueous humor transport. Specifically, the conventional OF ($C_{\text{trab}}$) is a composite of the intrinsic tissue permeability ($K_{\text{TM}}$, corresponding to the micro-scale parameter $u$) and the macroscopic geometric efficiency of the outflow pathway ($G \equiv A/L$, corresponding to the structural parameter $\gamma$). By defining $C_{\text{trab}}$ as the ratio of total aqueous flow ($Q_{\text{TM}}$) to the pressure drop ($\Delta P$) under Darcy's Law, we obtain the multiplicative composite:
\begin{equation}
	\label{eq:ctrab_composite}
	C_{\text{trab}} = Q_{\text{TM}} / \Delta P = K_{\text{TM}} \cdot G \cdot \mu^{-1}
\end{equation}
where $\mu$ is the dynamic viscosity of aqueous humor, and $G$ (units: m) represents the effective hydraulic cross-sectional area per unit path length. This maps exactly onto the abstract forward operator: the micro-scale operator $\mathcal{F}_{\text{micro}}(u, \gamma)$ yields $C_{\text{trab}}$, while the macro-scale operator $\mathcal{F}_{\text{macro}}$ corresponds to the Goldmann equation:
\begin{equation}
	\label{eq:goldmann}
	\text{IOP} = (Q_{\text{AH}} - F_u) / C_{\text{trab}} + \text{EVP}
\end{equation}
\begin{algorithm}[!t]
	 \SingleSpacedXII 
	\caption{Estimation Framework for 
		MHIP in Transport Systems}
	\label{alg:generalized_framework}
	\begin{enumerate}
		
		\item \textbf{Input:} Scalar macroscopic 
		observation $y^{\text{obs}}$, contextual 
		system metadata $c$.
		
		\item \textbf{Output:} Sampled latent 
		system inputs ($\mathbf{x}$) alongside 
		the resulting empirical push-forward 
		distributions for effective intrinsic 
		parameters ($\hat{u}_{\text{eff}}$), 
		effective structural parameters 
		($\hat{\gamma}_{\text{eff}}$), and 
		derived macroscopic quantities 
		(QoI, $\hat{q}$).
		
		\vspace{0.3em}
		\item[] \textbf{PHASE 1: Epistemic 
			Sampling (Latent System Inputs)}
		
		\item Generate $N$ plausible latent 
		system input samples 
		$\{\mathbf{x}_i\}_{i=1}^N$ from 
		population-level conditional 
		distributions 
		$\mathbb{P}(\mathbf{x} \mid c)$, 
		subject to domain-specific 
		biophysical plausibility constraints.
		
		\item For each state $i$, construct 
		the base feature vector 
		$\mathbf{X}_i = \{y^{\text{obs}}, 
		\mathbf{x}_i, c\}$.
		
		\vspace{0.3em}
		\item[] \textbf{PHASE 2: Two-Stage 
			Physics-Guided Evaluation}
		
		\item \textbf{for} $i = 1 \to N$ 
		\textbf{do}
		
		\item \hspace{1.5em} \textbf{[Stage 1: 
			Constrained Inverse Surrogate]}
		
		\item \hspace{1.5em} Predict effective 
		intrinsic parameter: 
		$\hat{u}_{\text{eff},i} \leftarrow 
		\mathcal{S}_1(\mathbf{X}_i)$.
		
		\vspace{0.2em}
		\item \hspace{1.5em} Construct augmented 
		feature vector: 
		$\mathbf{Z}_i = \{\hat{u}_{\text{eff},i}, 
		\mathbf{X}_i\}$.
		
		\vspace{0.2em}
		\item \hspace{1.5em} \textbf{[Stage 2: 
			Empirical Structural Calibration]}
		
		\item \hspace{1.5em} Predict effective 
		structural parameter: 
		$\hat{\gamma}_{\text{eff},i} \leftarrow 
		\mathcal{C}(\mathbf{Z}_i)$.
		
		\item \textbf{end for}
		
		\vspace{0.3em}
		\item[] \textbf{PHASE 3: Distribution 
			Synthesis and Uncertainty 
			Quantification}
		
		\item For each state $i$, derive 
		secondary macroscopic quantities 
		of interest ($\hat{q}_i$) by 
		evaluating the known closed-form 
		analytical physical laws governing 
		the system: 
		$\hat{q}_i \leftarrow 
		f(\hat{u}_{\text{eff},i}, 
		\hat{\gamma}_{\text{eff},i}, c)$, 
		where $f$ is fully specified by 
		the governing physics and requires 
		no additional learning.
		
		\item Aggregate the discrete ensemble 
		predictions $\{\hat{u}_{\text{eff},i}, 
		\hat{\gamma}_{\text{eff},i}, 
		\hat{q}_i\}_{i=1}^N$ to construct 
		marginal empirical push-forward 
		distributions.
		
		\item Extract predictive uncertainty 
		metrics (e.g., Coefficient of 
		Variation, empirical percentile 
		intervals) from the synthesized 
		distributions to explicitly quantify 
		the variance propagated from the 
		epistemic latency of $\mathbf{x}$.
		
		\item \textbf{return} Sampled latent 
		inputs $\{\mathbf{x}_i\}_{i=1}^N$, 
		the resulting empirical push-forward 
		distributions $\{\hat{u}_{\text{eff}}, 
		\hat{\gamma}_{\text{eff}}, \hat{q}\}$, 
		and derived uncertainty metrics 
		characterizing the variance propagated 
		from the epistemic latency of 
		$\mathbf{x}$.
		
	\end{enumerate}
	\end{algorithm}
which combines $C_{\text{trab}}$ with the latent boundary conditions $\mathbf{x} = \{Q_{\text{AH}}, F_u, \text{EVP}\}$ to produce the macroscopic observation. Critically, $\mathcal{F}_{\text{micro}}(u, \gamma)$ is physically independent of $\mathbf{x}$, as $C_{\text{trab}}$ is an intrinsic structural property determined solely by $K_{\text{TM}}$ and $G$. The boundary conditions $\mathbf{x}$ enter exclusively at the macroscopic level ($\mathcal{F}_{\text{macro}}$). Consequently, to learn an approximate left-inverse, the surrogate $\mathcal{S}_1$ must ingest both $y^{\text{obs}}$ and $\mathbf{x}$ to disentangle the macroscopic confounding; traversing the hierarchy in reverse requires knowledge of all variables that participated in the forward direction.

Substituting Equation~\ref{eq:ctrab_composite} into Equation~\ref{eq:goldmann} confirms the structural non-identifiability identified in Section~\ref{sec:statistical&algebraic}: for any fixed IOP under fixed boundary conditions $\mathbf{x}$, the product $K_{\text{TM}} \cdot G$ satisfies:
\begin{equation}
	\label{eq:ridge}
	K_{\text{TM}} \cdot G = \mu(Q_{\text{AH}} - F_u) / (\text{IOP} - \text{EVP}) = \text{constant}
\end{equation}
This identity formally proves the existence of the non-identifiable hyperbolic ridge, providing a mathematical rationale for our physics-guided architectural decomposition.

The \textit{Constrained Inverse Surrogate} ($\mathcal{S}_1$, Stage 1) is trained on high-fidelity FE simulations under a fixed, idealized geometry $\gamma^{\text{ideal}}$ to predict the functional permeability biomarker $\hat{K}_{\text{TM}}$. This parameter inherently contains the systematic structural bias ($\mathcal{D}_{\text{mismatch}}$) arising from the manifold mismatch between $\gamma^{\text{ideal}}$ and $\gamma^{\text{true}}$. 

Because the system remains structurally unidentifiable from a single scalar observation (Equation~\ref{eq:ridge}), the \textit{Empirical Structural Calibration Operator} ($\mathcal{C}$, Stage 2) does not attempt an impossible deterministic 3D reconstruction. Instead, functioning as an \textit{empirical prior conditioning step}, it maps the biased micro-scale estimate to the most probable macroscopic target ($\hat{G}$, units: m), conditional on demographic metadata (Age) and population-anchored clinical distributions. As formalized in Equation~\ref{eq:error_decomp}, Stage 2 is supervised against targets computed from $\mathcal{S}_1$'s actual biased predictions. This mathematically encodes structural error-compensation into the architecture, calibrating the biased micro-scale estimate against clinical $\hat{C}_{\text{trab}}$ targets. Training this operator requires a massive, clinically-anchored dataset, generated via PCDS ( Section~\ref{sec:pcds}).

\subsection{Computational Model Development for Stage 1 Data Generation}
The ground-truth dataset for the Stage 1 Constrained Inverse Surrogate ($\mathcal{S}_1$) was generated via a 2D axisymmetric FE model parameterized by anatomically-averaged dimensions \citep{RN35,RN36,RN37,RN38,RN39}. We use this idealized 2D representation because Stage 2 calibrates the remaining geometric effects empirically. We did not compare it against patient-specific 3D models. 

The model simulated coupled heat transfer and fluid dynamics (Pennes bioheat, Navier-Stokes, and Darcy's Law) \citep{RN40,RN41,RN37,RN38,RN39,RN42,RN43,RN44,RN45,RN46,RN47}. All inputs, including boundary conditions, static thermophysical properties, and age-stratified population distributions for key hydrodynamic parameters ($Q_{\text{AH}}$, $F_u$, EVP), were derived from clinical and experimental data \citep{RN45,RN48,RN49,RN39,RN10,RN13,RN50,RN51,RN52,RN53,RN54}. A constrained rejection sampling protocol ensured physiological plausibility. The system was solved using commercial FE software (COMSOL Multiphysics®, v6.3, COMSOL AB, Stockholm, Sweden), full implementation is detailed in \textbf{Online Supplement Note 2}.

\subsection{Formulation of the Supervised Learning Tasks}
Rather than embedding physical constraints via PDE residuals during model optimization, our framework enforces physical consistency architecturally. We formulate the estimation as two sequentially linked supervised regression tasks: Stage 1 isolates the effective intrinsic permeability ($\hat{K}_{\text{TM}}$), and Stage 2 computes the macroscopic geometric calibration factor ($\hat{G}$). The first stage learns the constrained inverse surrogate, $\mathcal{S}_1$, that predicts the log-transformed effective permeability from a vector of known hydrodynamic state variables, $\mathbf{X}$:
\begin{equation}
	\log_{10}(\hat{K}_{\text{TM}}) \approx \mathcal{S}_1(\mathbf{X})
	\label{eq:stage1_map}
\end{equation}
\begin{equation}
	\mathbf{X} = \{\text{IOP}, Q_{\text{AH}}, F_u, \text{EVP}, \text{AgeGroup}\}
	\label{eq:stage1_features}
\end{equation} 
where AgeGroup is a categorical variable determining the age-specific FE model geometry and physiological boundaries. The second stage learns the empirical structural calibration operator, $\mathcal{C}$, that predicts the log-transformed anatomical geometry factor from an augmented feature vector, $\mathbf{Z}$:
\begin{equation}
	\log_{10}(\hat{G}) \approx \mathcal{C}(\mathbf{Z})
	\label{eq:stage2_map}
\end{equation}
\begin{equation}
	\mathbf{Z} = \{\hat{K}_{\text{TM}}, \text{IOP}, Q_{\text{AH}}, F_u, \text{EVP}, \text{Age}\}
	\label{eq:stage2_features}
\end{equation}
where $\hat{K}_{\text{TM}}$ is the biased effective parameter estimated by $\mathcal{S}_1$, and Age is the patient's continuous numerical age. For both tasks, we employed XGBoost regressors \citep{RN55}, selected over evaluated baselines for their optimal balance of accuracy and model parsimony. Models were trained on an 80/20 split with hyperparameters optimized via 5-fold cross-validated randomized search. Predictive fidelity was evaluated on held-out test data, and the physical plausibility of the learned mappings was verified using SHapley Additive exPlanations (SHAP) \citep{RN56} (see \textbf{Online Supplement Note 5, Table S9, and Figure S6} for full details).

\subsection{Physics-Calibrated Data Scaling (PCDS) for Stage 2 Training}
\label{sec:pcds}
Training the Stage 2 Empirical Structural Calibration Operator ($\mathcal{C}$) requires a massive, clinically-anchored dataset. We did not generate this corpus with 3D FE simulations. At an estimated $\sim$4 hours per 3D run, 120,000 such runs would be computationally prohibitive. To circumvent this, we introduce PCDS. Grounded in multi-fidelity delta learning \citep{Peherstorfer2018}, PCDS calibrates an efficient analytical surrogate against the high-fidelity 2D FE model. First, a synthetic population ($N = 120,000$) was sampled from clinical fluorophotometry distributions \citep{RN11, RN12, RN13, RN21}. For each patient, an initial IOP was calculated using the Goldmann equation with explicit clinical measurement noise ($\epsilon_{\text{noise}}$):
\begin{equation}
	\text{IOP}_{\text{Goldmann}} = (Q_{\text{AH}} - F_u) / C_{\text{trab}} + \text{EVP} + \epsilon_{\text{noise}}
	\label{eq:goldmann_noise}
\end{equation}

Injecting $\epsilon_{\text{noise}}$ provides a critical regularization mechanism \citep{Nelsen2025}. A true inverse map is defined exclusively on the noiseless forward operator's exact range ($\text{Ran}(\mathcal{F})$). Because real-world tonometry measurements are inherently corrupted by noise, they fall outside this idealized manifold ($y^{\text{obs}} \notin \text{Ran}(\mathcal{F})$), causing severe mathematical instability in inverse operators trained solely on pristine physics data. Explicitly injecting $\epsilon_{\text{noise}}$ forces $\mathcal{C}$ to approximate a stable \textit{extension} of the true inverse map, ensuring robust processing of perturbed, off-manifold clinical data. Following noise injection, a systematic, pressure-dependent discrepancy between the analytical surrogate and the high-fidelity FE model was quantified ($P < 10^{-14}$). This bias was mitigated using a dynamic linear correction function (\textbf{Online Supplement Note 4}):
\begin{equation}
	\text{Bias}(\text{IOP}_{\text{Goldmann}}) = 2.654 - 0.233 \times \text{IOP}_{\text{Goldmann}}
	\label{eq:bias_correction}
\end{equation}

A first-order linear formulation was deliberately retained for pre-processing parsimony; the subsequent highly non-linear Stage 2 XGBoost ensemble natively absorbs any residual higher-order structural non-linearities. Furthermore, because vertical scatter around this regression represents the intentionally injected aleatoric noise ($\epsilon_{\text{noise}}$), the constrained goodness-of-fit ($R^2_{adj} = 0.119$) acts as a mathematical safeguard against overfitting to random clinical error.

The calibrated IOP values ($\text{IOP}_{\text{calibrated}} = \text{IOP}_{\text{Goldmann}} + \text{Bias}(\text{IOP}_{\text{Goldmann}})$) were then combined with the predicted $\hat{K}_{\text{TM}}$ from Stage 1 to generate the final augmented feature set ($\mathbf{Z}$) for all 120,000 entries. The full PCDS dataset was generated in approximately eight hours. This makes structural calibration computationally tractable.

\subsection{Framework Deployment and Probabilistic Profiling}
During deployment, the framework generates a comprehensive, patient-specific hydrodynamic profile from an individual's age (acting as the contextual metadata, $c$) and a single routine clinical measurement ($y^{\text{obs}} = \text{IOP}$). To manage the epistemic uncertainty of the unmeasured variables, the framework employs a constrained Monte Carlo simulation. Specifically, it generates $N = 1,000$ plausible latent system input vectors ($\mathbf{x}$) by sampling from age-stratified population distributions for aqueous inflow, uveoscleral outflow, and episcleral venous pressure ($Q_{\text{AH}}, F_u, \text{EVP}$). A rejection sampling step enforces strict domain-specific biophysical plausibility (e.g., $Q_{\text{AH}} > F_u$), and sensitivity analyses show that the outputs remain stable under large shifts in these assumed input distributions (\textbf{Online Supplement Note 3}).

These 1,000 sampled states are then sequentially evaluated by the trained two-stage architecture. The Constrained Inverse Surrogate ($\mathcal{S}_1$) first predicts the distribution of the effective intrinsic permeability ($\hat{K}_{\text{TM}}$). This output is subsequently augmented with the clinical inputs and passed to the Empirical Structural Calibration Operator ($\mathcal{C}$) to predict the macroscopic geometric factor ($\hat{G}$). From these paired outputs, the secondary macroscopic quantities of interest, the clinical OF ($\hat{C}_{\text{trab}}$) and the effective pore diameter ($\hat{D}_p$), are analytically derived using Darcy's Law and the Kozeny-Carman equation. By aggregating these discrete ensemble predictions, the framework constructs marginal empirical push-forward distributions for the complete biophysical state. The final synthesized output (visualized in Figure~\ref{fig:radar_plots}) provides not only the median point-estimates but also explicitly quantifies the predictive uncertainty (e.g., via the Coefficient of Variation) propagated from the epistemic latency of the unmeasured system inputs ($\mathbf{x}$). This probabilistic profile is ultimately normalized against a reference population and contextualized with data-driven risk thresholds to aid clinical interpretation without collapsing to a non-physical mean.

\subsection{Clinical Validation and Performance Metrics}
\label{sec:validation_methods}

To assess clinical generalizability, we curated two clinical datasets: a paired-eye steroid-challenge cohort ($n=19$) evaluated via a direct-measurement tonometer \citep{RN33}, and a heterogeneous meta-cohort of 27 clinical groups across 22 publications \citep{RN11,RN12,RN13,RN14,RN15,RN16,RN17,RN18,RN19,RN20,RN21,RN22,RN23,RN24,RN25,RN26,RN27,RN28,RN29,RN30,RN31,RN32}. Only cohort summary statistics (Age, IOP) served as model inputs. Measured OF was never a model input. Four of the 27 cohorts did not report measured OF, so OF comparisons use 23 cohorts. The Bland-Altman analysis also excludes the \textit{ex vivo} perfusion cohort, which leaves 22 cohorts. This validation is not fully independent of model development. Four of the meta-cohort studies also supplied the distributions used to generate PCDS training data \citep{RN11, RN12, RN13, RN21}. Four also supplied the distributions used to set the risk thresholds \citep{RN11,RN17,RN12,RN14}. Data-driven clinical risk thresholds (Normal, Borderline, Compromised) were established \textit{in silico} by mapping population distributions \citep{RN11,RN17,RN12,RN14} through our operators to extract the 25th and 75th percentiles of the resulting empirical push-forward distributions for $\hat{K}_{\text{TM}}$ (\textbf{Online Supplement Note 8}). Performance was evaluated via Bland-Altman analysis ($\hat{C}_{\text{trab}}$), Spearman's rank-correlation ($\hat{K}_{\text{TM}}$), and non-parametric rank tests.

To interpret the inferred functional permeability ($\hat{K}_{\text{TM}}$) microstructurally, we analytically derived an effective pore diameter ($\hat{D}_p$) using the Kozeny-Carman equation \citep{RN57}:
	\begin{equation}
		\hat{K}_{\text{TM}} = \hat{D}_p^2 \epsilon^3 / [k (1 - \epsilon)^2]
		\label{eq:kozeny_carman}
	\end{equation}
where $\epsilon$ is TM porosity and $k=150$ is the Kozeny constant. We assign $\epsilon$ by age group: 0.123 (young), 0.0915 (middle-aged) and 0.060 (older). These values are based on juxtacanalicular porosity estimates \citep{RN9, RN46}. Their assignment to age groups is our modeling choice. Because this model inherently assumes a bundle-of-tubes pore geometry, $\hat{D}_p$ represents a functionally equivalent hydraulic diameter rather than a direct morphological measurement of the trabecular microarchitecture.

\section{Results}
We first establish the predictive fidelity of the framework's core predictive operators on synthetic data before presenting a multi-faceted clinical validation of the derived macroscopic biomarkers.

\subsection{Predictive Fidelity of the Two-Stage Architecture}
Performance on held-out test sets confirmed the predictive operators learned physically plausible relationships rather than arbitrary data-fitting (\textbf{Online Supplement Note 5, Figure S6}). The Constrained Inverse Surrogate ($\mathcal{S}_1$) achieved an $R^2$ of 0.77 against ground-truth $\log_{10}(K_{\text{TM}})$ targets. Notably, residual analysis revealed distinct heteroscedasticity, with predictive error increasing for higher $K_{\text{TM}}$ values. This is consistent with physical ill-conditioning at lower IOPs (high permeability): as the pressure gradient approaches zero, the non-linear inverse approximation becomes inherently hyper-sensitive. Tiny fluctuations in boundary conditions are amplified into predictive variance, rendering a perfect $R^2$ practically unattainable. In contrast, the Empirical Structural Calibration Operator ($\mathcal{C}$) achieved an $R^2$ of 0.79 on log-transformed $\hat{G}$-factor targets with a strictly homoscedastic residual plot, indicating effective mitigation of the systematic geometric bias. Furthermore, SHAP analysis confirmed $\mathcal{C}$ correctly prioritized the biased Stage 1 estimate ($\hat{K}_{\text{TM}}$), consistent with the decomposed architecture compensating for the manifold mismatch ($\mathcal{D}_{\text{mismatch}}$) within expected algorithmic error bounds.

\subsection{Clinical Validation and Risk Stratification}
Having established the fidelity of the predictive operators on synthetic data, we assessed their \textit{in vivo} performance. Because true intrinsic permeability ($K_{\text{TM}}$) is unmeasurable \textit{in vivo}, the Stage 1 prediction ($\hat{K}_{\text{TM}}$) acts as an effective parameter containing a systematic geometric bias from the idealized reference manifold. Consequently, rather than assessing absolute error, we validate this functional biomarker via rank correlation against its closest clinical surrogate, measured OF.
\begin{figure*}[!t]
	\centering
	\caption[Validation of derived effective TM permeability]{
		\normalfont
		Validation of derived effective TM permeability ($\hat{K}_{\text{TM}}$). 
		\textbf{a--c}, Rank correlation of median $\hat{K}_{\text{TM}}$ empirical push-forward distributions against measured OF for (\textbf{a}) a heterogeneous meta-cohort ($n{=}23$), and a paired-eye cohort ($n{=}19$) of (\textbf{b}) untreated and (\textbf{c}) steroid-treated eyes. 
		\textbf{d, e}, Biomarker sensitivity in paired eyes, demonstrating steroid-induced reduction of (\textbf{d}) predicted $\hat{K}_{\text{TM}}$ and (\textbf{e}) analytically derived effective pore diameter ($\hat{D}_p$). Box plots: median, IQR, 1.5x IQR whiskers.
	}
	\label{fig:4}
	\includegraphics[width=\textwidth]{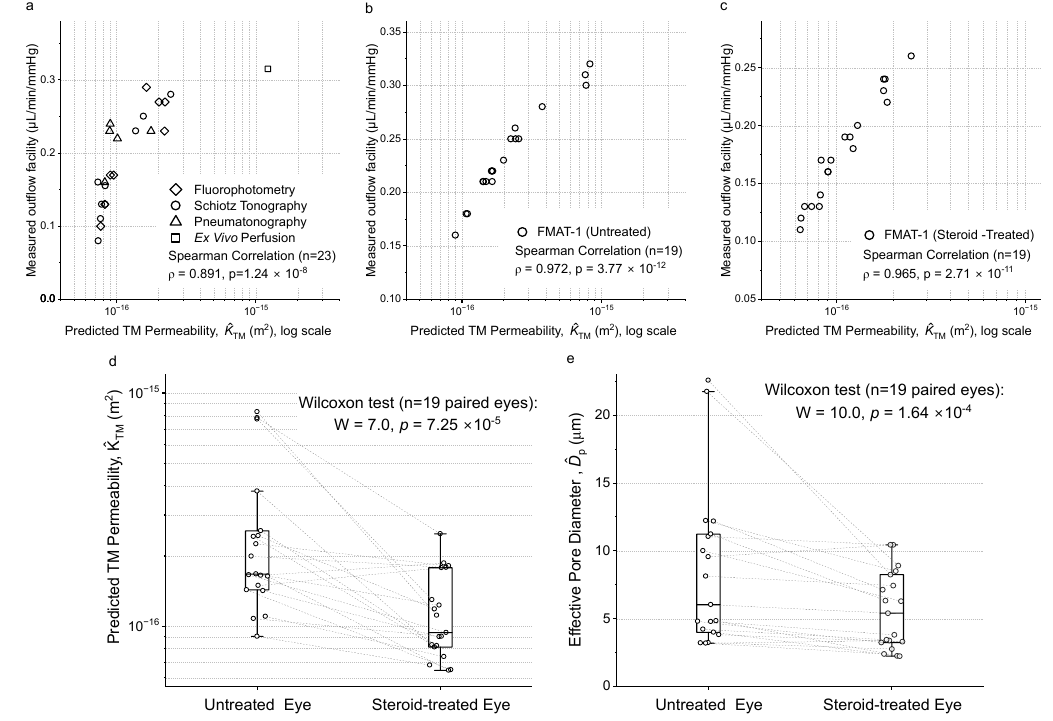}
\end{figure*}
\subsubsection{Relation Between the Effective Permeability Biomarker ($\hat{K}_{\text{TM}}$) and Outflow Facility}
While Equation~\ref{eq:ctrab_composite} establishes $C_{\text{trab}}$ as the multiplicative composite of $K_{\text{TM}}$ and $G$, their contributions to pathological outflow resistance are heavily skewed. We assume that variation in macroscopic geometric efficiency ($G \equiv A/L$) is small compared with the pathological loss of intrinsic permeability ($K_{\text{TM}}$). Under this assumption, $K_{\text{TM}}$ is the dominant source of variation in open-angle glaucoma. Because our Stage 1 surrogate isolates $\hat{K}_{\text{TM}}$ under a fixed reference manifold ($\gamma^{\text{ideal}}$), it is then expected to be monotonic with respect to $C_{\text{trab}}$. Testing this rank correlation against clinical data checks whether our non-invasive $\hat{K}_{\text{TM}}$ estimate captures true physiological variance. Across a heterogeneous validation set ($n=23$ cohorts), the medians of the empirical push-forward distributions for $\hat{K}_{\text{TM}}$ demonstrated a strong rank correlation with clinically measured OF values (Spearman's $\rho = 0.891$; $P = 1.24 \times 10^{-8}$; Figure~\ref{fig:4}a). This correlation approached unity in a homogeneous, paired-eye cohort ($n=19$) \citep{RN33} for both untreated ($\rho = 0.972$; $P = 3.77 \times 10^{-12}$; Figure~\ref{fig:4}b) and steroid-treated fellow eyes ($\rho = 0.965$; $P = 2.71 \times 10^{-11}$; Figure~\ref{fig:4}c). Together, these findings indicate that despite its structural bias, the derived $\hat{K}_{\text{TM}}$ estimate tracks outflow function closely by rank. The FMAT-1 correlations should be read with the caveat in Section~\ref{sec:ctrab_agreement}.
\subsubsection{Sensitivity of $\hat{K}_{\text{TM}}$ to Steroid-Induced Pathology}
The biomarker’s sensitivity to pathological change was tested using the paired-eye, steroid-challenge model ($n=19$). The framework predicted a marked and statistically significant decrease in $\hat{K}_{\text{TM}}$ in the steroid-treated eyes compared to their untreated fellow eyes ($P = 7.25 \times 10^{-5}$, Wilcoxon signed-rank test; Figure~\ref{fig:4}d). To provide a microstructural correlate, $\hat{K}_{\text{TM}}$ was translated into an effective hydraulic pore diameter ($\hat{D}_p$), which revealed a consistent and significant reduction in pore size following steroid treatment ($P = 1.64 \times 10^{-4}$; Figure~\ref{fig:4}e). The median $\hat{D}_p$ of the untreated eyes is $\sim$7~$\mu$m. This value is a model-derived equivalent diameter, not a measured pore size. We therefore do not treat it as external validation. It is reported as an interpretable summary of $\hat{K}_{\text{TM}}$ for the juxtacanalicular tissue, the primary site of outflow resistance \citep{RN8}.

\subsubsection{The Functional Biomarker ($\hat{K}_{\text{TM}}$) for Clinical Risk Stratification}
\begin{figure}[!t] 
	\centering
	\caption[Diagnostic performance and risk stratification]{ 
		\normalfont
		Diagnostic performance and risk stratification using the derived functional biomarker ($\hat{K}_{\text{TM}}$). 
		\textbf{a}, Distribution of the median predicted $\hat{K}_{\text{TM}}$ for 27 clinical cohorts, grouped by their ground-truth diagnostic status (Normal, Borderline, Compromised). Box plots show median, interquartile range, and 1.5x IQR whiskers. 
		\textbf{b}, Confusion matrix of the classification performance for the 27 cohorts (92.6\% accuracy, 25 of 27).
		See Online Supplement Note 6 for $\hat{C}_{\text{trab}}$ based risk-stratification comparisons.
	}
	\label{fig:risk_strat}
	\includegraphics[width=\columnwidth]{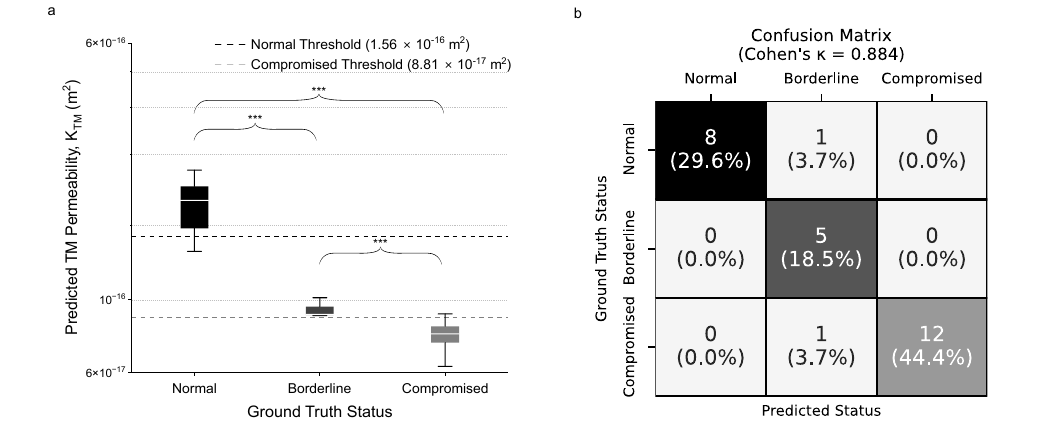} 
\end{figure}

To evaluate the clinical utility of the functional $\hat{K}_{\text{TM}}$ biomarker for granular risk stratification, we classified the heterogeneous set of 27 clinical cohorts according to their established diagnostic status, assigned via a pre-specified rule-based system (\textbf{Online Supplement Note 10}). For each cohort, a representative $\hat{K}_{\text{TM}}$ value was extracted from the median of its empirical push-forward distribution and classified into Normal, Borderline, or Compromised categories using our \textit{in silico} derived thresholds. The analysis confirmed that the cohort-level $\hat{K}_{\text{TM}}$ estimates for the three ground-truth risk groups were statistically distinct (Kruskal-Wallis test, $P = 2.20 \times 10^{-5}$; Figure~\ref{fig:risk_strat}a), with post-hoc pairwise Mann-Whitney U tests confirming significant separation between all groups ($P < 0.001$). The framework demonstrated high diagnostic accuracy, correctly classifying 25 of the 27 cohorts (92.6\%) and achieving a Cohen’s $\kappa$ of 0.884, indicating strong agreement with ground-truth clinical status (Figure~\ref{fig:risk_strat}b). This robust classification holds even at the population level across simulated \textit{in silico} cohorts (\textbf{Online Supplement Note 8 and Figure S7}).

\subsubsection{Agreement Between Estimated ($\hat{C}_{\text{trab}}$) and Measured Outflow Facility}
\label{sec:ctrab_agreement}
\begin{figure*}[t!]
	\centering
	\caption[Validation of derived conventional OF]{ 
		\normalfont
		Validation of derived conventional OF ($\hat{C}_{\text{trab}}$) against clinical measurements. 
		\textbf{a--c}, Bland-Altman analysis comparing the median $\hat{C}_{\text{trab}}$ empirical push-forward predictions against clinical data for (\textbf{a}) a heterogeneous cohort, and a paired-eye cohort of (\textbf{b}) untreated and (\textbf{c}) steroid-treated eyes. 
		\textbf{d}, Full empirical push-forward distributions of the difference between the 1,000 Monte Carlo $\hat{C}_{\text{trab}}$ estimates and measured OF per paired-eye patient. Violin plots contrast untreated (white) and steroid-treated (black) conditions, demonstrating individualized epistemic uncertainty.
	}
	\label{fig:6}
	
	\includegraphics[width=\textwidth]{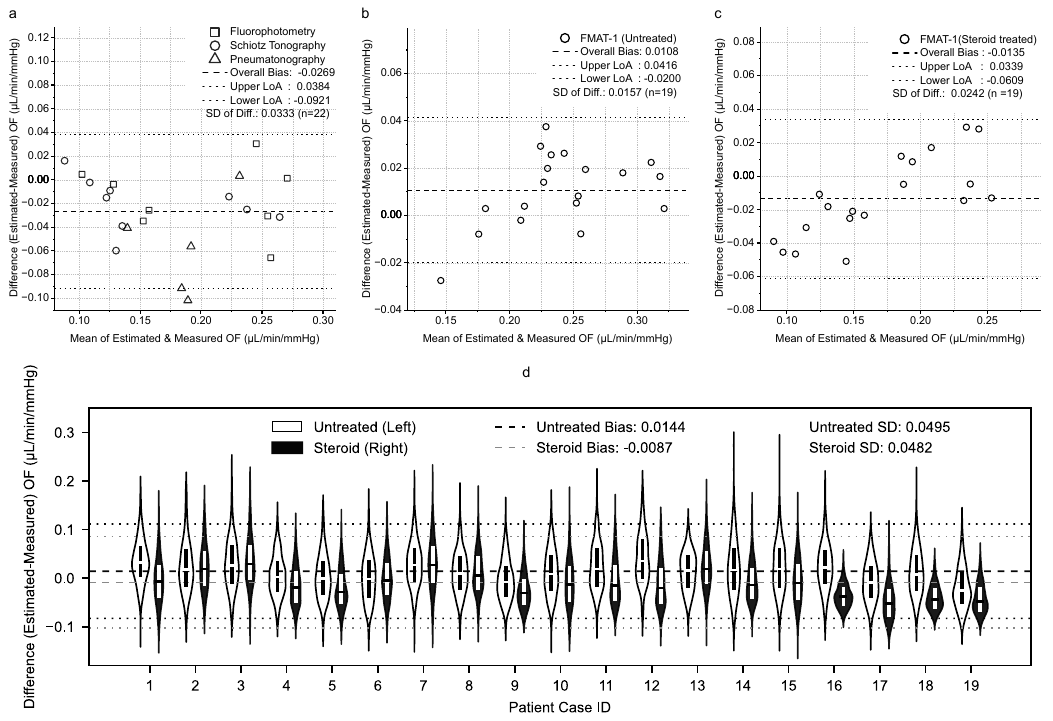}
	
\end{figure*}

The framework's clinical translatability relies on the Empirical Structural Calibration Operator (Stage 2) successfully reconciling the idealized biophysics with macroscopic reality. This was tested by comparing the derived parameter $\hat{C}_{\text{trab}}$ with measured OF. Generalizability was first assessed against the heterogeneous validation set ($n=22$ cohorts). The framework's cohort-level estimates maintained low overall bias and reasonable agreement with measured OF summary data from diverse measurement techniques (mean difference = -0.027~$\mu$L/min/mmHg; 95\% LoA[-0.092, 0.038]; Figure~\ref{fig:6}a). We then compared the estimates with FMAT-1 tonography in the paired-eye cohort. Agreement showed near-zero bias and narrow limits of agreement for both untreated (mean difference = 0.011~$\mu$L/min/mmHg; 95\% LoA [-0.020, 0.042]; Figure~\ref{fig:6}b) and steroid-treated eyes (mean difference = -0.013~$\mu$L/min/mmHg; 95\% LoA[-0.061, 0.034]; Figure~\ref{fig:6}c). The ICC for untreated eyes was 0.918 (95\% CI 0.690 to 0.970; \textbf{Online Supplement Table S13}). These FMAT-1 outflow values should not be read as fully independent measurements. In this cohort, IOP strongly predicted FMAT-1 outflow facility in treated eyes \citep{RN33}. Our estimates are also driven by IOP. Part of this agreement may therefore reflect the shared IOP signal.

Finally, to characterize the framework's predictive uncertainty, we visualized the full empirical push-forward distribution for each FMAT-1 patient (Figure~\ref{fig:6}d). This showed low bias on a per-patient basis. The distributions were wider for clinically atypical cases. This spread reflects the assumed variability of the unmeasured inputs. It is not a calibrated measure of prediction error.

\subsection{Integrated Ocular Hydrodynamic Profiling: Clinical Case Studies}

\begin{figure*}[!t]
	\centering
	\caption[Patient-specific hydrodynamic profiles from case studies]{ 
		\normalfont
		Patient-specific hydrodynamic profiles from clinical case studies. 
		Each radar plot shows the full empirical push-forward distribution (faint lines), median estimate (solid line), and population median (dashed hexagon) for all hydrodynamic parameters. 
		\textbf{a--c}, Paired-eye analyses from the FMAT-1 steroid-challenge cohort \citep{RN33}, comparing untreated (blue) and steroid-treated (red) eyes.
	}
	\label{fig:radar_plots}
	\includegraphics[width=\textwidth]{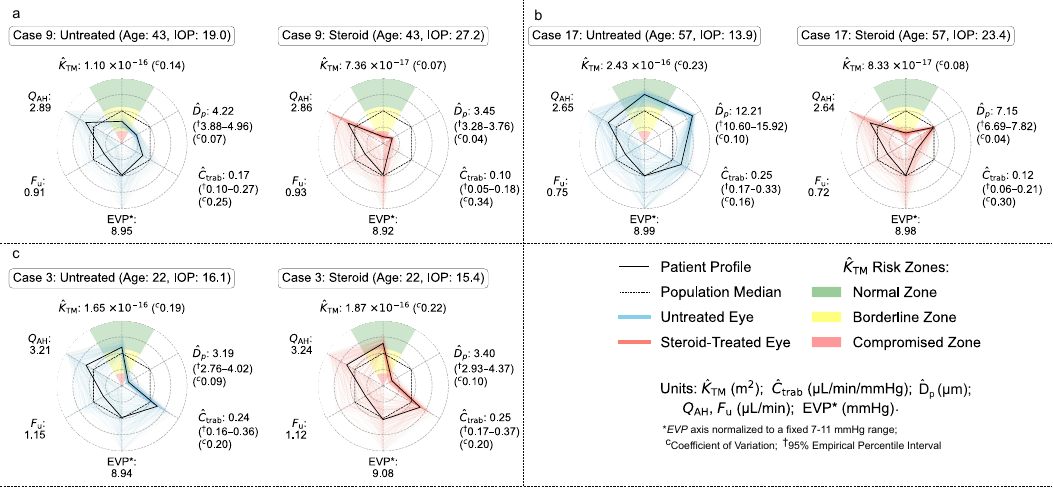}	
\end{figure*}

The framework synthesizes all operator outputs into an integrated, patient-specific profile, computationally deconstructing a single IOP measurement. Paired-eye analyses from the FMAT-1 cohort (Figure~\ref{fig:radar_plots}) illustrate this capability by successfully isolating the primary hydrodynamic mechanism of IOP elevation: a significant reduction in the derived functional biomarkers ($\hat{K}_{\text{TM}}$ and $\hat{C}_{\text{trab}}$). By projecting the complete parameter space, these radar plots capture the complex interplay of ocular hydrodynamics, illustrating how shifts in the biophysical root cause influence the broader system. Routine clinical measurements do not provide this joint view. A complementary six-visit longitudinal profile of a single OAG patient over four years \citep{RN34} is detailed in \textbf{Online Supplement Note 9 (Figure S8)}  illustrating its possible use for longitudinal monitoring.

This profiling moves beyond static risk assessment by quantifying predictive uncertainty alongside median estimates. Plausible hydrodynamic states are visualized via the overlaid Monte Carlo ensemble (faint lines), with explicit uncertainty bounds numerically reported via 95\% empirical percentile intervals and the Coefficient of Variation (CV). Notably, the functional biomarker $\hat{K}_{\text{TM}}$ frequently exhibits a lower CV than the composite clinical metric $\hat{C}_{\text{trab}}$, particularly at elevated IOP. The CV summarizes the spread propagated from the assumed input distributions. It has not been validated against observed prediction errors, so we do not use it as a reliability index. The lower relative spread suggests that $\hat{K}_{\text{TM}}$ may be a more stable target than the composite metric in high-IOP patients. Prospective data are needed before the framework is used to track progression or guide therapy.

\section{Discussion}
\label{sec:discussion}

This study introduces a comprehensive computational framework that, to our knowledge, presents a novel formal solution strategy for the MHIP. Using the clinical challenge of glaucoma as a proof-of-concept, we address a class of severely underdetermined problems characterized by extreme data scarcity and structural non-identifiability. By leveraging a physics-guided architectural decomposition, our approach computationally deconstructs a single macroscopic measurement (IOP) into a probabilistic profile of its underlying biophysical root causes. In doing so, this work provides a direct, mechanistic alternative to simple lumped-parameter statistical models, addressing a fundamental gap in the computational development of precision models from sparse data.

The framework’s success rests on two methodological pillars. First, our two-stage architecture explicitly separates constrained biophysical approximation ($\mathcal{S}_1$) from empirical structural calibration ($\mathcal{C}$). Our ablation study suggests that this decomposition reduces the predictive collapse seen in monolithic models under multiplicatively entangled physics. Second, PCDS makes this architecture computationally tractable. It generates large, physically consistent training datasets without running high-fidelity 3D simulations. Clinically, this allows the framework to translate a systematically biased, intermediate parameter ($\hat{K}_{\text{TM}}$) into a low-bias estimate of the clinical gold standard ($\hat{C}_{\text{trab}}$). Furthermore, by mapping the epistemic latency of unmeasured system inputs via Monte Carlo sampling, the framework generates empirical push-forward distributions, giving a probabilistic physiological profile rather than a single point estimate.

These capabilities stem from deliberate methodological trade-offs prioritizing macroscopic clinical accuracy over microscopic morphological purity. By modeling the TM as a homogenous medium and forcing Stage 1 onto a fixed, idealized 2D manifold ($\gamma^{\text{ideal}}$), the derived $\hat{K}_{\text{TM}}$ acts as a systematically biased bulk parameter. Similarly, the structural calibration operator ($\mathcal{C}$) functions as an empirical bridge to \textit{in vivo} 3D complexity, rendering the inferred $\hat{G}$-factor an effective calibration tool rather than a true anatomical reconstruction. However, this explicit architectural trade-off, accepting a systematically biased intermediate biomarker ($\hat{K}_{\text{TM}}$) in order to empirically calibrate it ($\hat{G}$) into a low-bias clinical prediction ($\hat{C}_{\text{trab}}$), is precisely the mathematical mechanism that bypasses the structural non-identifiability of the MHIP. The retrospective validation is encouraging, but the framework's diagnostic utility requires prospective clinical evaluation.

\paragraph{Limitations.} This study has several limitations. First, once IOP and the sampled inputs are fixed, the Goldmann equation (Equation~\ref{eq:goldmann}) gives $C_{\text{trab}}$ in closed form. We have not compared Stage 2 against this analytical estimate or against other statistical baselines. Second, the empirical push-forward distributions are conditional on the assumed input distributions and on the trained surrogates. The sensitivity analyses test stability under changes to these assumptions. They do not show that the distributions are calibrated patient-specific uncertainty. Third, the validation is retrospective and not fully independent. Some meta-cohort studies also informed the training distributions and the risk thresholds. The FMAT-1 outflow values are closely tied to IOP. A simple rule based on IOP alone may also classify the 27 cohorts well. We have not tested such a baseline. Fourth, the fixed 2D reference geometry was not compared with patient-specific 3D models. The inferred $\hat{G}$ is a calibration factor, not a recovered patient anatomy. Addressing these points is the focus of ongoing work.
\bibliographystyle{plainnat}
\bibliography{references}

\clearpage
\includepdf[pages=-]{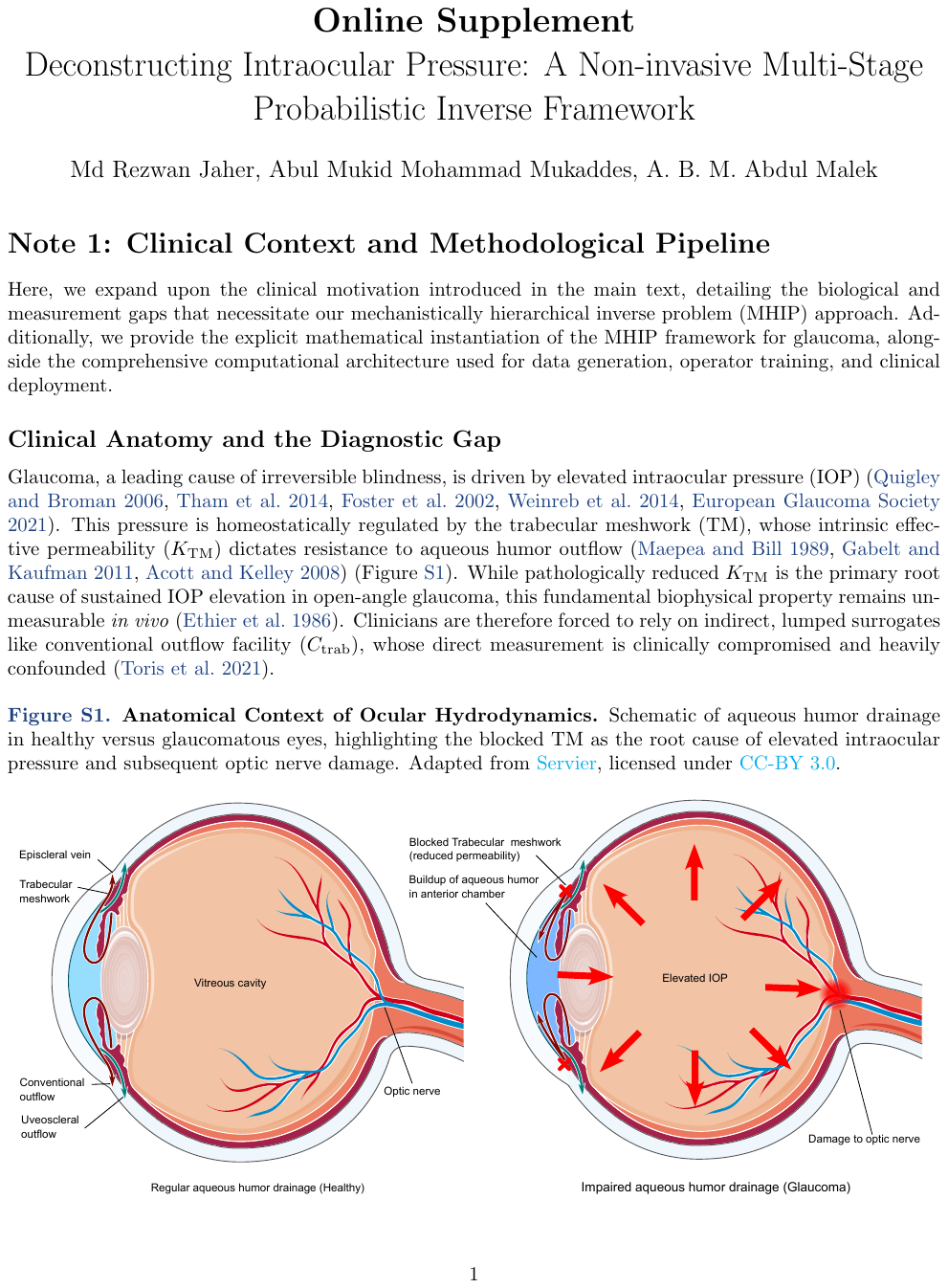}

\end{document}